\documentclass[conference]{IEEEtran}
\IEEEoverridecommandlockouts

\usepackage{cite}
\usepackage{amsmath,amssymb,amsfonts}
\usepackage{graphicx}
\usepackage{textcomp}
\usepackage{xcolor}
\usepackage{booktabs}
\usepackage{url}

\usepackage{tikz}
\usetikzlibrary{positioning,arrows.meta,fit,calc,backgrounds}

\definecolor{qubitblue}{RGB}{70,130,180}
\definecolor{aigreen}{RGB}{50,150,50}
\definecolor{controlred}{RGB}{220,20,60}
\definecolor{grayc}{RGB}{120,120,120}

\begin{document}

\title{The Third Restructuring of Software Form: From the Three-Tier Architecture to Storage, Models, and Agents}

\author{\IEEEauthorblockN{Wei Lin\textsuperscript{1}, Tao Zhou\textsuperscript{1}, Zhaofei Xie\textsuperscript{1}, Changgui Hong\textsuperscript{1}}
\IEEEauthorblockA{\textsuperscript{1}Nanjing Liancheng Intelligent Technology Group, Nanjing, China\\
Email: \{linwei, zhoutao, xiezhaofei, hongchanggui\}@chinaliancheng.com}
}

\maketitle


\begin{abstract}
Software form has undergone two paradigm shifts since its inception: Software 1.0, in which instructions determine behavior, and Software 2.0, in which data determines behavior (machine learning). This paper argues that a third shift---Software 3.0, in which context and reasoning determine behavior---is now underway, and contends that its terminal form converges to three elements: a \emph{generalized database} (the unified abstraction of all persistent state and memory), a \emph{large model} (the intelligence core that performs reasoning and generation), and an \emph{agent} (the execution loop connecting the first two). The core argument is as follows: in the traditional three-tier architecture, the user-interface layer will be absorbed by the model's ability to generate interfaces on demand, the business-logic layer will be re-partitioned along ``expressibility $\times$ criticality'' into model reasoning and storage constraints (with residual deterministic logic retained as tools), and only the data layer will be elevated into the sole persistent infrastructure. We formalize this convergence thesis, present a minimal reference architecture, report evidence from real prototypes and a live model, and systematically analyze both the conditions under which it holds and the boundaries where it fails---determinism, cost, security, and verifiability delimit the thesis's domain of applicability. We argue that the thesis holds in task domains that are expressible, verifiable, externally stateful, and tool-complete, and that it will reshape the roles of developers, the database industry, and the software-engineering discipline.
\end{abstract}

\section{Introduction}
\label{sec:intro}

Software is the medium by which humans prescribe machine behavior, and its form has never been constant: it restructures itself whenever the cost of expressing behavior drops. Marc Andreessen's claim that ``software is eating the world'' presupposes that software is cheap enough to build and easy enough to reuse~\cite{andreessen2011software}. When large language models (LLMs) reduce the cost of translating natural language into executable behavior to an unprecedented low, the \emph{form of software itself} ceases to be a given and becomes a variable worth re-examining.

This paper pursues a question that appears radical yet already shows abundant signs: \emph{if interfaces can be generated instantaneously by a model, and business rules can be reasoned about instantaneously by a model, what remains of traditional software?} Our answer: only three things remain---state (storage), intelligence (model), and execution (agent).

This thesis does not arise from a vacuum; it surfaces simultaneously across several independent research strands. Andrej Karpathy's ``LLM OS'' casts the large model as a kernel and the database as a file system~\cite{karpathy2017software20,karpathy2023llmos}. Work such as DBOS argues for making the database, rather than the operating system, the foundation of distributed applications~\cite{li2022dbos}. Agent research---ReAct, Toolformer, AutoGPT, Voyager---demonstrates that models can autonomously complete multi-step tasks through think--act--observe loops~\cite{yao2022react,schick2023toolformer,autogpt2023,wang2023voyager}. Work on retrieval-augmented generation (RAG) and MemGPT shows that external storage and memory are the key to transcending the context window and acquiring long-term state~\cite{lewis2020rag,packer2023memgpt}. These strands are independent, yet they point toward the same convergence.

Our contributions are as follows:

\begin{itemize}
\item \textbf{Thesis formalization}: we elevate ``software = storage + model + agent'' from a slogan to a discussable, testable proposition, with precise boundaries and relations among the three elements (Section~\ref{sec:formal}).
\item \textbf{Collapse mechanism}: we systematically argue why each layer of the three-tier architecture is either absorbed or elevated (Section~\ref{sec:collapse}).
\item \textbf{Minimal reference architecture}: we give an end-to-end architecture showing how the three elements compose into a complete, working software system (Section~\ref{sec:arch}).
\item \textbf{Boundary analysis}: we state the conditions under which the thesis holds and the counterexamples where it fails, so the vision does not degenerate into a slogan (Sections~\ref{sec:conditions}--\ref{sec:boundaries}).
\item \textbf{Implications}: we discuss what the thesis means for developers, the database industry, the software-engineering discipline, and governance (Section~\ref{sec:discussion}).
\end{itemize}

\section{Background and Related Work}
\label{sec:background}

This section reviews the four strands that support the thesis and locates our contribution relative to them.

\subsection{``Software 2.0'' and ``Software 3.0''}

In 2017, Karpathy introduced ``Software 2.0'': software behavior is no longer specified by explicit code but is implicit in neural-network weights trained on data~\cite{karpathy2017software20}. This insight captures the first shift, from ``program-specified behavior'' to ``data-determined behavior.'' Along the same logic, the community has begun to sketch ``Software 3.0'': a further decision factor---\emph{context and reasoning}---is layered on top of Software 2.0; a model is no longer a static, trained function but a system that dynamically decides behavior at runtime based on prompts, tool feedback, and external memory. Our thesis can be read as the most radical version of Software 3.0: once context and reasoning dominate behavior, the only durable part of software is storage.

\subsection{The LLM Operating System}

Since the Transformer architecture~\cite{vaswani2017attention} and GPT-style scaled pretraining~\cite{brown2020gpt3} established large models as general intelligence cores, Karpathy's 2023 ``LLM OS'' analogy casts the large model as the kernel (CPU/RAM), external tools as peripherals (I/O), and the database/file system as persistent storage~\cite{karpathy2023llmos}. The analogy is evocative but remains metaphorical. This paper asks the follow-up question: if the metaphor is taken literally, what is the \emph{minimal} composition of a system with a model as its core and storage as its foundation? Our answer: storage, model, and agent together suffice to close the loop.

\subsection{Agents and Tool Use}

The breakthrough that lets models ``not only speak but also act'' comes from two lines of work. The first couples reasoning with action: building on the step-by-step reasoning capability established by chain-of-thought prompting~\cite{wei2022cot}, ReAct proposes an interleaved think--act--observe paradigm in which a model invokes external tools during reasoning and learns from feedback~\cite{yao2022react}, and Toolformer shows that models can learn to call APIs through self-supervised learning~\cite{schick2023toolformer}. The second closes the autonomous task loop: AutoGPT/BabyAGI demonstrate an automatic goal--plan--execute cycle~\cite{autogpt2023}, and Voyager goes further by letting an agent accumulate reusable capabilities through a skill library in Minecraft~\cite{wang2023voyager}. Together these works establish the agent as an independent \emph{execution loop}---neither the model itself, nor hand-written glue code, but the institutional carrier of the plan--memory--tool-use dynamics.

\subsection{The Convergence of Databases and AI}

A fourth independent strand is the convergence of databases and AI. On the query side, text-to-SQL translates natural language into structured queries, sharply lowering the barrier to data access. On the storage side, vector databases make semantic similarity a first-class capability that directly serves RAG~\cite{lewis2020rag}. On the memory side, MemGPT proposes a hierarchical memory architecture that lets a model manage context the way an operating system manages memory, with an external database serving as the model's ``long-term memory''~\cite{packer2023memgpt}. DBOS approaches the same destination from the opposite direction, arguing that the database---not the OS---should be the foundation of distributed applications, with transactions, scheduling, and logging all provided as database primitives~\cite{li2022dbos}. The shared implication of this strand is that \emph{the database is being elevated from a passive storage appendage into an active infrastructure}.

\subsection{Gaps in Prior Work and Our Positioning}

These strands each make progress, but share a common blind spot: they study ``how a model becomes the kernel,'' ``how an agent becomes the execution loop,'' and ``how a database becomes the foundation'' \emph{separately}, yet few works unify all three into a single convergence thesis at the \emph{macro level of software form}, and answer the question ``what gets replaced, what does not, and under what conditions.'' This paper fills that gap with an integrated treatment of the thesis, its argument, an architecture, and its boundaries.

\begin{figure}[t]
\centering
\begin{tikzpicture}[
  strand/.style={draw, rounded corners=2pt, fill=grayc!10, text width=2.6cm, align=center, font=\footnotesize\sffamily, inner sep=3pt},
  elem/.style={draw, rounded corners=2pt, fill=qubitblue!18, text width=1.9cm, align=center, font=\footnotesize\sffamily, inner sep=4pt},
  arr/.style={->,>=Stealth,semithick}
]
\node[strand] (p1) {Software 2.0 / 3.0};
\node[strand, below=0.25cm of p1] (p2) {LLM OS};
\node[strand, below=0.25cm of p2] (p3) {Agents \& Tool Use};
\node[strand, below=0.25cm of p3] (p4) {Database $\times$ AI};

\node[elem, right=1.3cm of p1] (M) {Large Model};
\node[elem, right=1.3cm of p3] (A) {Agent};
\node[elem, right=1.3cm of p4] (D) {Generalized Database};

\draw[arr] (p1) -- (M);
\draw[arr] (p2) -- (M);
\draw[arr] (p2.east) to[out=0,in=180] (D.west);
\draw[arr] (p3) -- (A);
\draw[arr] (p3.east) to[out=0,in=180] (M.south);
\draw[arr] (p4) -- (D);
\draw[arr] (p4.east) to[out=0,in=180] (A.south);
\end{tikzpicture}
\caption{Convergence of four independent research strands onto the three elements; each strand contributes to one or more elements, jointly pointing toward the storage--model--agent convergence.}
\label{fig:convergence}
\end{figure}
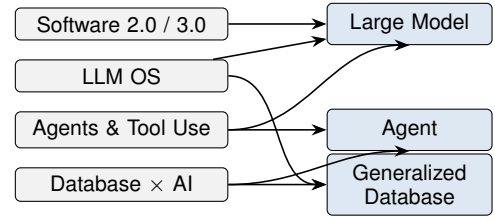

\subsection{A Skeptical View and Our Response}

A substantial body of work cautions against overstating LLM capability. Surveys of code hallucination document that models generate plausible but incorrect outputs with no correctness guarantee~\cite{gao2025hallucination}, and empirical studies of industry needs report that reliability and explainability remain the top concerns that current academic approaches fail to address~\cite{yu2025aligning}. These findings are often read as evidence against agentic software.

This paper does not dispute them; rather, the convergence thesis is built to be \emph{compatible} with them. We do not claim the model is reliable---we claim the model should not have to be. By confining the model to the expressible, non-critical tier and guaranteeing the critical tier through deterministic storage constraints (Section~\ref{sec:threesplit}), the thesis limits the damage of hallucination to the region where verification can catch it (Section~\ref{sec:boundaries}). The skeptical view therefore strengthens, rather than weakens, the central role the thesis assigns to the storage layer.

\section{Formalizing the Thesis}
\label{sec:formal}

\subsection{Statement}

Let a software system $S$ be the composition of the three traditional tiers:
\begin{equation}
S = (U, L, D)
\end{equation}
where $U$ is the user-interface layer, $L$ is the business-logic layer, and $D$ is the data layer. Our central thesis is:

\noindent\textbf{Convergence Thesis}: within the task domain satisfying the conditions of Section~\ref{sec:conditions}, $U$ is absorbed by the model's on-demand generation, $L$ is absorbed by the model's reasoning and tool use, and the software form converges to:
\begin{equation}
S' = (\mathcal{D}, \mathcal{M}, \mathcal{A})
\end{equation}
where $\mathcal{D}$ is the \emph{generalized database}---the unified storage abstraction of all persistent state, constraints, memory, and knowledge (a single semantic layer over heterogeneous relational, vector, graph, key-value, and object stores); $\mathcal{M}$ is the \emph{large model}---the intelligence core carrying understanding, reasoning, generation, and decision-making; and $\mathcal{A}$ is the \emph{agent}---an execution loop structured as plan--memory--tool-use, acting as the dynamic connector between model and storage.

\subsection{Precise Definitions of the Three Elements}

To keep the thesis from degenerating into a slogan, we define the three concepts precisely:

\begin{enumerate}
\item \textbf{The generalized database} is not a single database product, but a unified abstraction of the \emph{functional role} of ``persistent state.'' It encompasses structured relational data, semi-structured documents, unstructured vectors and objects, together with the constraints (schemas, integrity rules, permissions) and version history that describe them. Its defining property is that it is the \emph{only} part of the system possessing persistence, auditability, and transactionality.
\item \textbf{The large model} is the carrier of the functional role of ``intelligence.'' We presuppose no specific model, but require two indispensable capabilities: \emph{reasoning} (making decisions under a given context) and \emph{generation} (translating decisions into executable actions or interfaces).
\item \textbf{The agent} is the carrier of the functional role of ``execution loop.'' Its essence is a \emph{closed loop}: perceive context $\rightarrow$ plan $\rightarrow$ invoke tools (read/write $\mathcal{D}$) $\rightarrow$ observe results $\rightarrow$ update memory $\rightarrow$ continue. The agent stitches the stateless reasoning of the model and the stateful storage together into a whole that keeps working over time.
\end{enumerate}

The relation among the three can be summarized as: \emph{storage is the software's ``past'' (memory and state), the model is its ``present'' (reasoning and decision), and the agent is its ``future'' (advancing past and present into the next moment).}

\section{Argument: Why the Three-Tier Architecture Collapses}
\label{sec:collapse}

This section argues, layer by layer, why each tier is absorbed or elevated.

\subsection{The UI Layer Dissolves: Interfaces Generated on Demand}

The traditional UI layer exists to present business capability in a human-perceivable form. But a UI is, at bottom, a translation from state to presentation. When a model can perform this translation reliably, statically pre-built interfaces are no longer a necessity---an interface can be generated at each interaction from the \textbf{current state, user intent, and device context}. This trend is already visible in ``generative UI'' and ``chat-as-interface.''

But ``dissolution'' must be qualified precisely, lest we repeat the mistake of treating reliability as a free premise. A UI is not monolithic; it splits into two layers: a \textbf{deterministic projection layer} and a \textbf{generative decoration layer}. The former carries information that ``must be presented correctly''---account balances, contract amounts, compliance disclosures, medical warnings, accessibility semantics---whose probabilistic mis-generation constitutes a substantive safety/compliance failure, not a cosmetic blemish; it must be obtained as a \emph{deterministic projection} of stored state (state-driven UI). The latter---layout, wording, interaction rhythm, and other ``expressible and non-critical'' presentation---may be delegated to on-demand generation by the model.

In other words, the UI layer's dissolution obeys the \emph{same law} as the business-logic layer's differentiation: the expressible and non-critical part goes to the model, and the must-be-correct part anchors on storage. The UI layer therefore does not vanish; it is restructured from ``a pre-hard-coded whole'' into ``a deterministic projection of storage plus a generative decoration by the model.'' This restructuring also exposes the traditional usability costs of a UI---consistent mental models, branding, accessibility---which belong to the part the deterministic projection layer must retain, not the latitude of the generative layer.

\subsection{The Business-Logic Layer Dissolves: Reasoning and Tool Use Replace Hard-Coded Rules}
\label{sec:threesplit}

The business-logic layer ($L$) is the most expensive and the most perishable part of software: it consists of large amounts of branching, rules, and glue code, and it corrodes as requirements drift. When a model can reason about business rules from context and trigger side effects through tool calls, $L$ is re-partitioned---\emph{not into a clean dichotomy}, but along the two axes of ``expressibility $\times$ criticality'' into three kinds:

\begin{enumerate}
\item \textbf{Expressible and non-critical}: rules that can be clearly stated in language and whose errors are tolerable, absorbed by the model's reasoning;
\item \textbf{Critical but declaratively expressible}: rules that map onto deterministic primitives such as uniqueness, foreign keys, CHECK constraints, triggers, and transactions, sunk into storage constraints on $\mathcal{D}$;
\item \textbf{Critical yet not declaratively expressible}: multi-step cross-system orchestration with external calls, temporal dependencies, and intricate exception branches---these exceed the expressive ceiling of declarative constraints and cannot be carried cleanly; this part \emph{does not vanish}, but survives as verified, deterministic code exposed to the agent as controlled tools.
\end{enumerate}

In short, \emph{business logic does not undergo a clean ``polarization'': it differentiates into three kinds, the third of which constitutes a crucial correction to the thesis's most naive reading}---not all logic can be absorbed by model or constraints, and the thesis's scope is precisely delimited by how small this third kind can be made. Table~\ref{tab:threesplit} grounds the three kinds in a concrete domain---intelligent production scheduling (APS)---which we carry through Section~\ref{sec:arch}.

\begin{table}[t]
\centering
\caption{The three-way differentiation of business logic, instantiated with rules from an intelligent production-scheduling (APS) system.}
\label{tab:threesplit}
\setlength{\tabcolsep}{4pt}
\begin{tabular}{p{2.0cm}p{3.4cm}p{2.2cm}}
\toprule
\textbf{Kind} & \textbf{Example scheduling rule} & \textbf{Lands in} \\
\midrule
Expressible, non-critical & ``Why is order \#A delayed?''; summarize today's plan & Model reasoning \\
Critical, declaratively expressible & One machine runs one operation at a time; an operation starts only after its predecessors; capacity is never exceeded & Storage constraints (uniqueness, precedence, CHECK) \\
Critical, not declaratively expressible & Minimize total tardiness / makespan; balance due dates against changeover cost & Deterministic solver tool (e.g., CP-SAT) \\
\bottomrule
\end{tabular}
\end{table}

\subsection{The Data Layer Rises: The Generalized Database Becomes the Sole Persistent State}

Once $U$ and $L$ retreat, $D$ becomes the only surviving durable layer---and its status rises, not falls. Three reasons. First, model reasoning is \emph{stateless}; long-term state must be externalized, and the database is the only reliable carrier. Second, the model's capability boundary is precisely determined by ``what state it can see,'' so the database becomes the ceiling of the model's ability. Third, under hard requirements of auditability, rollback, and transactionality, only the database can provide deterministic guarantees~\cite{hellerstein2007architecture}. The data layer thus rises from ``an appendage dominated by the logic layer'' to ``the foundation of the entire system''---mutually reinforcing DBOS's ``database-as-foundation'' argument~\cite{li2022dbos}.

\subsection{The Agent: The Execution Loop Connecting Model and Storage}

With $U$ and $L$ gone and $D$ elevated to the foundation, the system still needs a mechanism that connects the stateless model to stateful storage so that software can \emph{keep working} rather than merely answer once. That is the agent's role, $\mathcal{A}$. The agent is not a new layer conjured from nothing; it is the institutionalization of the plan--memory--tool-use loop. It decides what state to read next, which tool to call, what result to write back, and what to remember. In this sense, \textbf{the agent replaces precisely the ``main loop'' that was frozen into the control flow of traditional software.}

In summary, the collapse of the three-tier architecture can be expressed in one sentence: \emph{the UI regresses into a model output, the logic differentiates into model reasoning, storage constraints, and residual deterministic tools, the control flow is reconstructed as the agent's loop, and only the state survives---and is elevated.} Table~\ref{tab:collapse} summarizes the mapping.

\begin{table}[t]
\centering
\caption{From the three-tier architecture to three elements.}
\label{tab:collapse}
\setlength{\tabcolsep}{4pt}
\begin{tabular}{p{3.2cm}p{4.5cm}}
\toprule
\textbf{Traditional} & \textbf{Converged form} \\
\midrule
User interface layer & Generated on demand by the model \\
Business logic layer & Three-way split: model reasoning, storage constraints, residual deterministic tools \\
Data layer & Generalized database (sole persistent state, plus constraints and history) \\
Control flow (implicit) & Agent execution loop (plan--memory--tool) \\
\bottomrule
\end{tabular}
\end{table}

\section{A Reference Architecture: A Minimal Storage--Model--Agent System}
\label{sec:arch}

This section gives the thesis a concrete landing form, showing how the three elements compose into a complete, working system.

\subsection{Overall Architecture}

\begin{figure*}[t]
\centering
\begin{tikzpicture}[
  node distance=0.5cm and 0.6cm,
  every node/.style={font=\footnotesize\sffamily},
  box/.style={draw, rounded corners=1.5pt, fill=#1!22, align=center, inner sep=4pt, minimum height=0.7cm, minimum width=2.0cm},
  grp/.style={draw=black!35, rounded corners=3pt, inner xsep=8pt, inner ysep=10pt, fill=black!4},
  arr/.style={->,>=Stealth,semithick},
  dasharr/.style={->,>=Stealth,semithick,dashed}
]
\node[box=grayc] (P) {Planner};
\node[box=grayc, right=1.2cm of P] (M) {Memory};
\node[box=grayc, right=1.2cm of M] (T) {Tool\\Dispatcher};

\node[box=qubitblue, below=2.0cm of P] (LLM) {Large Language\\Model};

\node[box=aigreen, below=1.8cm of LLM] (R) {Relational};
\node[box=aigreen, right=0.6cm of R] (V) {Vector};
\node[box=aigreen, right=0.6cm of V] (G) {Graph};
\node[box=aigreen, right=0.6cm of G] (KV) {Object/KV};
\node[box=aigreen, below=0.5cm of V] (S) {Constraints / History};

\node[box=controlred, right=1.0cm of T] (W) {External\\World};

\begin{scope}[on background layer]
\node[grp, fit=(P)(M)(T), label={[font=\footnotesize\sffamily\bfseries]south:Agent Layer (Execution Loop)}] (Ag) {};
\node[grp, fit=(LLM), label={[font=\footnotesize\sffamily\bfseries]left:Model Layer}] (Md) {};
\node[grp, fit=(R)(V)(G)(KV)(S), label={[font=\footnotesize\sffamily\bfseries]left:Storage Layer}] (St) {};
\end{scope}

\draw[arr,<->] (P) -- (LLM);
\draw[arr] (P.north) to[bend left=25] (T.north);
\draw[arr] (P) -- (M) node[midway, below, font=\footnotesize\sffamily] {retrieve};
\draw[arr] (T) -- (M);
\draw[arr,<->] (T) -- (W);
\draw[arr] (T.south) -- (St.north) node[midway, right=6pt] {read / write};
\draw[arr] (M.south) to[out=-90,in=90] (V.north);
\draw[dasharr] (S.north) -- (V.south);
\end{tikzpicture}
\caption{A minimal reference architecture for a storage--model--agent system: a stateless model core, an agent execution loop, a heterogeneous generalized database, and the external world reached only through tools.}
\label{fig:architecture}
\end{figure*}
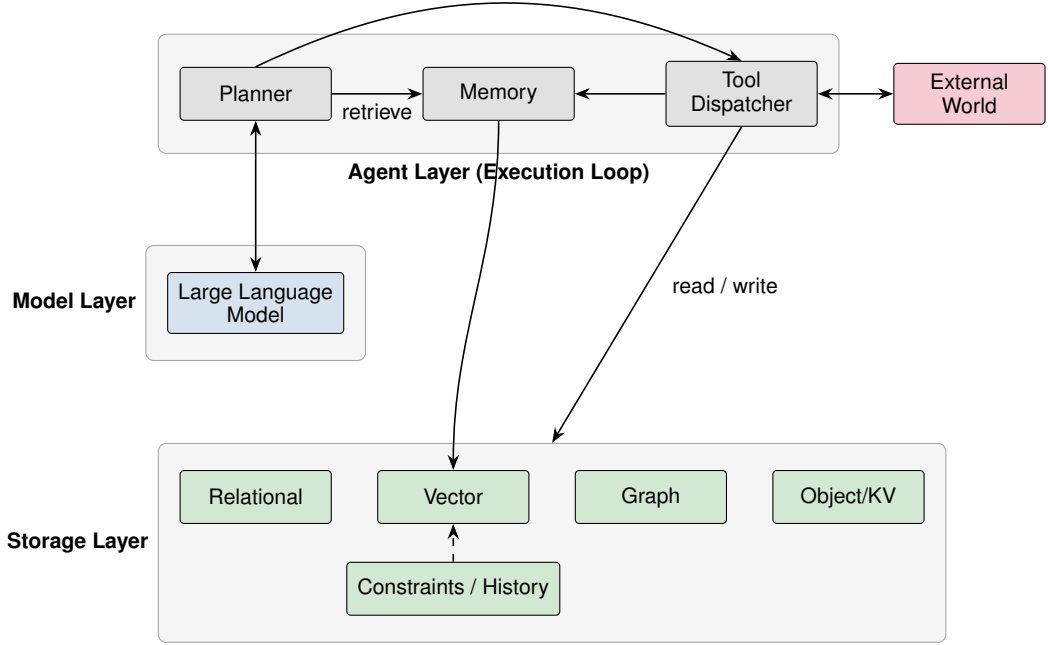

\subsection{Storage Layer: Unified Semantics over Heterogeneous Stores}

The storage layer is not ``one database,'' but a unified abstraction over heterogeneous relational, vector, graph, and object stores---continuing the ``one-size-fits-all no longer applies'' line of specialized engines~\cite{stonebraker2005onesizefitsall}---augmented with two kinds of semantics: (1) \textbf{constraints}---the parts of business logic that ``must be correct'' are sunk into deterministic storage-level constraints (uniqueness, foreign keys, transactions, stored procedures, triggers), guaranteed by the database engine rather than the model; and (2) \textbf{version and history}---the full evolution trace of state is preserved for auditing, rollback, and verifiability, the source of reliability and interpretability for long-running agents.

\subsection{Model Layer: A Stateless Intelligence Core}

The model layer does three things: it parses user intent and context into a \textbf{plan}; it decomposes the plan into \textbf{executable actions}; and it synthesizes raw tool results into a \textbf{user-comprehensible presentation}. The model itself remains stateless---all long-term state is hosted in the storage layer. This ``stateless core + stateful external memory'' division is a continuation, in a new era, of the classical database/stateless-service layering.

\subsection{Agent Layer: Plan--Memory--Tool}

The agent layer consists of three cooperating modules: the \textbf{Planner}, which decomposes a goal into steps, decides which tools to call and in what order, and revises the plan dynamically based on feedback; the \textbf{Memory}, which maintains short-term context and long-term memory (persistent knowledge written into the storage layer), transcending the model's context-window limit~\cite{packer2023memgpt}; and the \textbf{Tool Dispatcher}, which executes reads/writes to the storage layer and calls to the external world in a controlled manner---the sole exit through which the agent produces side effects.

\subsection{A Proof-of-Concept: Intelligent Production Scheduling}

We ground the architecture in an intelligent production-scheduling (APS) scenario---a long-lived, constraint-rich domain where the convergence thesis applies most directly. A conventional APS is a monolithic application coupling a scheduling interface, hard-coded dispatching rules, and a relational database. In the converged form it reduces to three elements.

\textbf{Storage layer.} Orders, operations, machines, and material inventories are persisted as state, together with the declarative constraints that must never be violated: one machine executes one operation at a time (an exclusion constraint), an operation starts only after all its predecessors finish (a precedence constraint), and a schedule must respect machine capacity and material availability (integrity constraints). These are enforced by the database engine, not by the model.

\textbf{Model layer.} The model interprets natural-language requests---``re-schedule shop floor A around the urgent order \#2047,'' or ``why is order \#1881 slipping?''---into a plan over the available tools, and renders the returned schedule and its rationale back into prose.

\textbf{Agent layer.} For a re-scheduling request, the Planner decomposes the goal into steps: read the affected orders and current assignments from the storage layer, invoke the \emph{solver tool} (a deterministic constraint/optimization routine, e.g., a CP-SAT solver) to compute a new feasible schedule under the stored constraints, write the resulting assignment back, and update Memory with the incident context. The Tool Dispatcher is the sole channel through which the solver and the shop-floor systems are invoked.

The three-way split of Section~\ref{sec:threesplit} is visible end to end: the expressible, non-critical parts---diagnosing a delay or summarizing a plan---are model reasoning; the critical, declaratively expressible parts---mutual exclusion, precedence, capacity---are storage constraints; and the critical, non-declarative part---the combinatorial search for a schedule minimizing tardiness---remains a deterministic tool. A query such as ``why is order \#1881 late?'' is answered entirely by reading state and reasoning over the stored precedence and capacity facts; a request to re-schedule is executed by the agent loop while the storage layer guarantees the result remains feasible.

\subsection{Preliminary Evidence from a Minimal Prototype}
\label{sec:proto}

To make the storage layer's role concrete, we implemented a minimal prototype in Python and SQLite: a job shop of 10 machines and 200 unit-time operations (four precedence chains per machine), in which machine exclusion is a \texttt{UNIQUE(machine, slot)} constraint and precedence is a \texttt{BEFORE INSERT} trigger that raises on violation. The planner is a deliberately noisy scheduler that perturbs a fraction $\varepsilon$ of its assignments to random machine--slot pairs, standing in for an imperfect model.

Table~\ref{tab:proto} reports the result. Regardless of the planner's error rate---up to 30\% of assignments perturbed---the persisted schedule remains feasible, because the storage constraints reject every violating proposal (16, 33, and 45 rejections at $\varepsilon=0.1$, $0.2$, and $0.3$, respectively). Enforcement costs about 1.1 ms per 200 operations in SQLite. The prototype is intentionally minimal---the ``model'' is simulated---but it demonstrates the thesis's central mechanism: \emph{correctness is guaranteed by the storage layer, independent of upstream reasoning quality}.

\begin{table}[t]
\centering
\caption{Storage-as-arbiter: the persisted schedule stays feasible regardless of the planner's error rate $\varepsilon$.}
\label{tab:proto}
\setlength{\tabcolsep}{4pt}
\begin{tabular}{cccccc}
\toprule
$\varepsilon$ & Proposals & Rejected & Persisted & Feasible & Time (ms) \\
\midrule
0.0 & 200 & 0 & 200 & yes & 1.23 \\
0.1 & 200 & 16 & 184 & yes & 1.14 \\
0.2 & 200 & 33 & 167 & yes & 1.11 \\
0.3 & 200 & 45 & 155 & yes & 1.08 \\
\bottomrule
\end{tabular}
\end{table}

The second experiment replaces the placeholder with a production solver. We model the combinatorial objective---minimizing makespan in a job shop, the kind of rule that is critical but not declaratively expressible---with OR-Tools CP-SAT, and run it on reproducible synthetic job-shop instances (fixed seeds). Table~\ref{tab:solver} reports solve time and makespan. The objective is solved to optimality in well under a second for most instances, but one hard $12\times12$ instance takes $9.67$~s---illustrating that this tier carries real, variable computational cost, and is therefore correctly isolated as a residual tool rather than fused into model reasoning or storage constraints.

\begin{table}[t]
\centering
\caption{The residual tool (OR-Tools CP-SAT) on reproducible synthetic job-shop instances.}
\label{tab:solver}
\setlength{\tabcolsep}{4pt}
\begin{tabular}{cccc}
\toprule
Instance & Makespan & Time (s) & Status \\
\midrule
6$\times$6 & 594 & 0.01 & optimal \\
8$\times$8 & 756 & 0.02 & optimal \\
10$\times$10 & 857 & 0.16 & optimal \\
12$\times$12 & 944 & 9.67 & optimal \\
15$\times$15 & 1143 & 0.56 & optimal \\
\bottomrule
\end{tabular}
\end{table}

\subsection{A Live Model: The Storage Layer Catches Real Hallucinations}
\label{sec:llm}

The preceding experiments simulated the model. To test the arbiter against a \emph{real} model, we asked a live LLM (Qwen-Plus, via a production API) to schedule the three-job, three-machine instance of Table~\ref{tab:llm} directly, without a solver tool, in twenty independent trials. The model returned well-formed schedules every time, yet \emph{zero} of them were feasible: all twenty violated machine exclusion, and eighteen also violated precedence, with apparent makespans of $7$--$10$ (mean $8.9$) that are meaningless precisely because they ignore machine contention---the true optimum is $9$. The storage layer rejected all twenty violating schedules, a 100\% catch rate, so the persisted state remained feasible throughout. This is the thesis's central mechanism at work: \emph{the model need not be reliable, because correctness is enforced by the storage layer.}

\begin{table}[t]
\centering
\caption{A live model (Qwen-Plus) scheduling a $3\times3$ job shop directly: 0\% feasible, 100\% caught by storage constraints.}
\label{tab:llm}
\setlength{\tabcolsep}{4pt}
\begin{tabular}{lc}
\toprule
Metric & Value \\
\midrule
Trials & 20 \\
Feasible schedules produced directly & 0 / 20 \\
Machine-exclusion violations & 20 / 20 \\
Precedence violations & 18 / 20 \\
Violating schedules rejected by storage & 20 / 20 \\
Apparent makespan, mean (optimum $= 9$) & 8.9 \\
Mean latency per call & 3.84 s \\
\bottomrule
\end{tabular}
\end{table}

The contrast with a tool-using agent completes the picture. When the same model is instructed to delegate to the deterministic solver tool rather than reason directly, it requests the tool in all ten trials and reports the correct makespan ($9$) and feasibility every time (Table~\ref{tab:tooluse}). The two experiments together show that the model's 0\% feasibility is not a limitation of the model per se, but of entrusting combinatorial correctness to reasoning: correctness comes from \emph{delegating to the deterministic tool} and is \emph{enforced by the storage layer}, exactly as the thesis's three-way split prescribes.

\begin{table}[t]
\centering
\caption{Tool-using agent (Qwen-Plus, 10 trials): 100\% correct via delegation to the solver tool.}
\label{tab:tooluse}
\setlength{\tabcolsep}{4pt}
\begin{tabular}{lc}
\toprule
Metric & Value \\
\midrule
Requested the solver tool & 10 / 10 \\
Reported correct makespan ($= 9$) & 10 / 10 \\
Stated the schedule is feasible & 10 / 10 \\
Mean latency per call & 4.82 s \\
\bottomrule
\end{tabular}
\end{table}

\subsection{Hand-Written vs. Declarative Enforcement}
\label{sec:handwritten}

The three-way split places the critical, declarative rules in the storage layer. To justify that placement, we compare two ways of expressing the same three scheduling constraints (machine exclusion, precedence, capacity): hand-written checks inline in application code, versus declarative constraints in the storage layer. The enforcement code is comparable in size, but the difference appears under fault injection: when a single check---the capacity rule---is forgotten in the hand-written version, over-capacity schedules \emph{leak through} (Table~\ref{tab:handwritten}); the same rule declared as a storage constraint is enforced regardless of what the application code does. This is not a claim that declarative constraints are novel---databases have long provided them---but that their role becomes \emph{critical} when the upstream logic is an unreliable model (Section~\ref{sec:llm}) rather than careful human code: centralized, unavoidable enforcement is the safety net that makes the converged form trustworthy.

\begin{table}[t]
\centering
\caption{Fault injection: a single forgotten check leaks in hand-written code; declarative storage constraints always enforce.}
\label{tab:handwritten}
\setlength{\tabcolsep}{4pt}
\begin{tabular}{lccc}
\toprule
Case & Hand-written & Hand-written (missing capacity) & Storage \\
\midrule
valid & caught & caught & caught \\
overlap & caught & caught & caught \\
precedence & caught & caught & caught \\
capacity & caught & \textbf{leaked} & caught \\
\bottomrule
\end{tabular}
\end{table}

\section{Conditions for the Thesis to Hold}
\label{sec:conditions}

The thesis does not hold unconditionally. This section states four necessary conditions; in task domains that fail them, the thesis's applicability sharply weakens (Section~\ref{sec:boundaries}).

\subsection{Expressibility and Verifiability of the Task}
\label{sec:express-verify}

The model's capability boundary is determined by two properties: whether the task can be \textbf{clearly expressed in language} (otherwise the model cannot reason about it), and whether the result can be \textbf{objectively verified} (otherwise the model's errors cannot be detected and corrected). When a task is both expressible and verifiable, model reasoning plus verification feedback forms a reliable feedback loop; otherwise the task reverts to traditional implementation. This condition directly echoes Sutton's ``bitter lesson''---whatever can be solved by computation and data will eventually be solved by general methods~\cite{sutton2019bitter}.

\subsection{Externalization of State}

The thesis requires that all long-term state be \textbf{externalized} into the storage layer. If the task's state is naturally embedded in the model (a one-shot, side-effect-free question answering), the storage-plus-agent loop is meaningless; only when software must maintain state \textbf{across time, sessions, and subjects} does the elevation of the storage layer make sense. The thesis therefore applies to \textbf{stateful, long-lived} software, not stateless one-off computation.

\subsection{Completeness of the Tool Boundary}
\label{sec:tool-boundary}

An agent can act on the world only through tools. The thesis therefore presupposes that the side effects required by the domain (reading/writing external systems, operating devices, calling services) can all be exposed to the agent as \textbf{controlled tools} with clear permission boundaries. But this condition contains an inherent dilemma:

\begin{enumerate}
\item \textbf{Completeness and closedness cannot both hold.} If the toolset is written down in advance, it is safe but incomplete---any unforeseen side effect becomes inexpressible, and the thesis reverts to ``still hand-writing code.'' If the toolset is open (letting the agent bootstrap new tools, as in Voyager's skill library~\cite{wang2023voyager}), it is complete but unbounded---``completeness'' degenerates into unlimited trust in the model's self-bootstrapping, which is no condition at all.
\item \textbf{A single tool's permission boundary cannot express multi-step composed side effects.} Even when every tool is individually well-scoped, the workflow-level side effects produced by an agent \emph{composing} multiple tools exceed the expressive power of a per-tool permission model---just as individually safe queries can jointly infer private information through their sequence.
\end{enumerate}

Together these yield a constructive corollary: since authorization cannot be fully closed at the individual-tool level, \textbf{the final enforcement point of authorization can only be the storage layer}---where every state change converges and can be constrained and audited (cf.\ Section~\ref{sec:governance}). The tool boundary is thus not a premise satisfiable once and for all, but an engineering constraint that the storage layer must continuously backstop.

\subsection{Economic Threshold}

Even if the first three conditions hold, the thesis's realization depends on \textbf{economics}. But the comparison needs clarifying: what actually faces off is not ``the latency/cost of one inference'' versus ``one execution of compiled logic''---these are \textbf{differently structured costs}. The cost of hand-writing $U$ and $L$ is dominated by \textbf{development and maintenance} (labor and requirement drift), with near-zero marginal runtime cost; the cost of the storage--model--agent form is dominated by \textbf{marginal runtime cost} (per-inference latency and billing), while its development and maintenance cost declines as model capability improves.

The true dividing line is therefore not ``call frequency'' but the \textbf{value density of a decision}---the ratio of the cost of a single inference to the value of the decision it produces. For low-value, high-concurrency decisions (each request worth a fraction of a cent), inference cost dominates and hand-written logic is more economical; for high-value, low-concurrency decisions (a single approval that averts a million-dollar loss), inference cost is a rounding error and the model form dominates. This boundary is also \textbf{dynamic and engineerable}: inference cost declines over the long run while maintenance cost does not, and caching, distillation, small models, and batching further compress runtime cost. ``Core high-frequency transactions'' is thus only a static snapshot; the true economic boundary is moving, over time, in the thesis's favor.

\subsection{Non-Triviality of the Thesis}

The conditions above may invite a criticism: if the thesis holds only in the domain where model reasoning and storage constraints happen to work, is it nearly tautological? We argue not, for two reasons. First, the conditions are not vacuous relaxations but are \textbf{jointly satisfiable by a real, identifiable, and continuously expanding class of software}---long-tail, natural-language-interfaced, stateful business workflows; Sutton's ``bitter lesson'' implies that, as model capability grows, this class only expands~\cite{sutton2019bitter}. Second, the thesis's substance is not the truism that ``models perform well in their home domain,'' but a falsifiable \textbf{architectural claim}: after the collapse of traditional software, the only remaining durable artifact is \emph{exactly} storage---not ``storage plus a thin layer of business logic.'' It is this ``exactly'' that gives the thesis predictive content against future evidence, rather than rendering it a tautology.

\section{Counterexamples and Boundaries: When the Thesis Fails}
\label{sec:boundaries}

The greatest danger of a vision paper is overpromising. This section enumerates four boundary classes where the thesis fails, and how they delimit its domain.

\subsection{Determinism and Correctness}

Relational databases have endured because they offer the deterministic guarantees of \textbf{transactions, types, and constraints} (ACID). Model reasoning is fundamentally a \emph{probabilistic} computation and cannot promise the same guarantees. For scenarios where errors are unacceptable---funds settlement, aerospace, medical dosing---business logic must be guaranteed by formally verifiable code or database constraints, not by model reasoning. This boundary means the thesis does not apply to \textbf{strongly deterministic, formally verified} core tasks; in such tasks $L$ does not vanish, but merely sinks into storage constraints or verified code.

\subsection{Performance, Latency, and Cost}

Model inference latency (seconds) and cost (per-call billing) far exceed the direct execution of compiled logic (nanoseconds, nearly free). In \textbf{high-frequency, low-latency, large-scale} scenarios (trade matching, real-time recommendation, network forwarding), freezing logic into dedicated implementations remains the only choice. In these scenarios the thesis fails, or degrades to a compromise where ``the model only generates code, but does not reason at runtime.''

\subsection{Security, Permissions, and Compliance}

Handing execution authority to an agent that ``decides its next move by reasoning'' opens a new attack surface and new compliance risks: prompt injection can hijack the agent's decisions; an ill-designed tool-permission boundary becomes a channel for privilege escalation (the composed-side-effect problem of Section~\ref{sec:tool-boundary}); and regulation (in finance and healthcare, for instance) often requires behavior to be \textbf{predictable, auditable, and attributable}, which conflicts with the opacity of model reasoning.

The hardest of these is the \textbf{attribution problem}, whose depth exceeds what ``adding audit logs'' can resolve: attribution demands an answer to ``who is responsible for a wrong decision,'' yet an agent's decision is an \emph{emergent} result of model reasoning, stored state, and historical tool feedback, with no single accountable subject to point to. Audit logs can tell us \emph{what happened}, not \emph{who is to blame}. Nor is ``human-in-the-loop'' an easy fallback: the \textbf{responsibility paradox} is that the model amplifies the capacity to act while the human's capacity to understand and to bear responsibility does not scale in kind---the human can neither grasp the model's full decision trace nor vouch for every action at scale.

This problem, in turn, reinforces one of the thesis's conclusions: since attribution must anchor on something \textbf{deterministic and auditable}, its only anchor can be the storage layer---who committed what state, when, and which constraint was violated; the model and the agent are not accountable, and the storage layer is the ultimate carrier of accountability. But this also exposes a \textbf{structural gap} in strongly regulated domains: the pure converged form leaves storage as the \emph{only} deterministic anchor, and the ``human'' component is absent. In such domains the thesis must therefore degrade to ``storage constraints + human-in-the-loop''---and how an unintelligible, hard-to-scale ``human'' is to backstop a superlinearly growing agent is a question the thesis has not answered.

\subsection{Verifiability and Hallucination}

Model hallucination means its output may be \emph{superficially plausible yet factually wrong}~\cite{gao2025hallucination}. When a task's result cannot be objectively verified, hallucination cannot be detected, and the thesis's feedback loop collapses. This again echoes Section~\ref{sec:express-verify}: verifiability is the thesis's lifeline. A notable corollary: \textbf{in the converged form, the database plays the role of an ``arbiter of consistency'' rather than an ``arbiter of fact''}---any model output that conflicts with already-stored state can be vetoed by the storage layer (a consistency constraint); but the database cannot adjudicate \emph{fabricated novel facts} that conflict with nothing already stored (e.g., inventing an entity that does not exist in the database), because such hallucinations violate no existing constraint. Correctness assurance for novel outputs must therefore come from external oracles, cross-checks, or humans-in-the-loop; the storage layer can only confine hallucination's harm to the ``expressible but not yet persisted'' segment.

\subsection{Software Categories That Will Not Be Replaced}

Synthesizing the above boundaries, we can state precisely which software the thesis will not replace---which, counterintuitively, strengthens the thesis's credibility:

\begin{table}[t]
\centering
\caption{Software categories not replaced by the convergence thesis, and the thesis's applicable form for each.}
\label{tab:notreplaced}
\begin{tabular}{p{2.4cm}p{2.7cm}p{2.4cm}}
\toprule
\textbf{Category} & \textbf{Why it is not replaced} & \textbf{Applicable form} \\
\midrule
Database / storage engine & Source of determinism, transactions, performance & Implementation of $\mathcal{D}$ \\
OS / runtime / compiler & Demand correctness, nanosecond performance & Invoked by $\mathcal{A}$ as tools \\
High-frequency transaction systems & Latency, determinism & Keep $L$; model offline only \\
Regulated / safety-critical systems & Accountability, auditability & Storage constraints + human-in-the-loop \\
Model training / inference frameworks & Carrier of $\mathcal{M}$ & The thesis's core, not replaceable \\
\bottomrule
\end{tabular}
\end{table}

\section{Discussion: Implications for Industry, Developers, and Governance}
\label{sec:discussion}

If the thesis holds within its domain, it carries several far-reaching implications.

\subsection{For Developers: From ``Writing Code'' to ``Defining State and Constraints''}

The developer's core work shifts from ``writing control-flow code that implements a specific behavior'' to \textbf{``defining the state model, designing constraints, encapsulating tools, and debugging the agent's behavior.''} This is not ``the disappearance of programmers'' but an \emph{upward shift} of the center of programming: from specifying ``how to do it'' line by line, to precisely describing ``what is correct and what is forbidden.'' This shift is already foreshadowed by ``vibe coding,'' but its mature form is \textbf{constraint-driven development}---developers invest effort in verifiable constraints rather than perishable glue logic~\cite{ge2025vibecoding}.

\subsection{For the Database Industry: From Passive Storage to Active Infrastructure}

The database rises from ``passive storage beneath applications'' to ``active infrastructure above them.'' This elevation is not idle speculation---its first half is already observable: vector retrieval has spawned an independent database category~\cite{lewis2020rag}, text-to-SQL has sharply lowered the barrier to structured-data access, and DBOS argues for rebuilding the foundation of distributed applications on database primitives~\cite{li2022dbos}. What this paper further predicts is its second half---the database assuming the new responsibilities of \textbf{``arbiter of consistency'' and ``governor of state''}: the authoritative source and final adjudicator of agent behavior.

This prediction is falsifiable: if, in a future ``agent + database'' form, state governance still requires a separate control plane outside the database, then this thesis's claim that ``storage is elevated into the sole foundation'' is correspondingly weakened. In other words, the value of the database industry will no longer be priced by ``access performance'' alone, but by ``how many deterministic semantic anchors''---constraints, audit, rollback, adjudication---it can supply.

\subsection{For the Software-Engineering Discipline: Repositioning Correctness Assurance}

The foundation of traditional software engineering is ``correctness by construction.'' Under the converged form, correctness assurance \textbf{polarizes into two poles}: \emph{probabilistic correctness} (model reasoning, secured by a verification-feedback loop) and \emph{deterministic correctness} (storage constraints and formal verification, secured by mathematics). The central question of software engineering thus partially shifts from ``how to write correct code'' to ``how to design constraints and verification such that a probabilistic system is, on the whole, trustworthy.''

\subsection{Risk and Governance}
\label{sec:governance}

The converged form brings new governance challenges: agent behavior must be \textbf{auditable} (fully recorded to the storage layer), \textbf{attributable} (clear responsibility for agent errors), and \textbf{limitable} (least-privilege tools and storage constraints). We argue that the healthy evolution of the converged form depends on folding ``governance'' itself into the storage layer's constraint and audit semantics---making governance part of the foundation rather than an after-the-fact patch.

\section{Conclusion}
\label{sec:conclusion}

This paper proposes and argues for a convergence thesis about software form: within task domains that are expressible, verifiable, externally stateful, tool-complete, and economically viable, the traditional three-tier architecture will collapse into three elements---\textbf{generalized database + large model + agent}---with the UI regressing into model output, the business logic differentiating into model reasoning, storage constraints, and residual deterministic tools, the control flow reconstructed as the agent's loop, and only the state surviving and being elevated into the foundation.

We have also delineated the thesis's boundaries: determinism and correctness, performance and cost, security and compliance, and verifiability and hallucination---these four constraint classes determine the thesis's domain, and in doing so illuminate \textbf{which software will not be replaced}. The value of this paper lies not in declaring a utopian ``end of software,'' but in offering a \textbf{testable framework}: one that points out the direction of software-form convergence while also marking the endpoint and the limits of that convergence path.

Future work can proceed in three directions: first, scale the experiments of Sections~\ref{sec:proto} and~\ref{sec:llm} to larger instances and a full tool-using agent, and compare against a hand-written baseline; second, formalize the mechanism of ``storage constraints as the arbiter of consistency over model outputs,'' and study whether it can provide a quantifiable correctness backstop for probabilistic reasoning; third, investigate the methodology and toolchain of ``constraint-driven development'' from a software-engineering perspective. We believe that whether the thesis is ultimately confirmed or refuted, the very act of posing it will deepen our understanding of what software truly is.

\section*{Artifact Availability}
\label{sec:artifacts}

The prototype and all experiment scripts are available at \url{https://github.com/kyloTyn/software-form-convergence}. The two live-model experiments call Qwen-Plus through the Alibaba DashScope compatible-mode API and require a key supplied via the \texttt{DASHSCOPE\_API\_KEY} environment variable; no key is stored in the repository. All other experiments are deterministic and depend only on Python and OR-Tools.

\bibliographystyle{ieeetr}
\bibliography{ref}

\end{document}